# Deep learning for determining a near-optimal topological design without any iteration

Yonggyun Yu, Taeil Hur, Jaeho Jung, and In Gwun Jang*

Yonggyun Yu

https://orcid.org/0000-0003-2863-9650

Korea Advanced Atomic Research Institute

Daejeon 34057, Republic of Korea

ygyu@kaeri.re.kr

Taeil Hur

https://orcid.org/0000-0002-1542-1057

Seoul 08005, Republic of Korea

taeilhur@gmail.com

Jaeho Jung

https://orcid.org/0000-0001-5203-5035

Korea Advanced Atomic Research Institute

Daejeon 34057, Republic of Korea

jungjaeho@kaeri.re.kr

In Gwun Jang

The Cho Chun Shik Graduate School of Green Transportation

Korea Advanced Institute of Science and Technology (KAIST)

291 Daehak-ro, Yuseong-gu, Daejeon 34141, Republic of Korea

*Corresponding author [+82-42-350-1262, igjang@kaist.edu]

**Abstract**

In this study, we propose a novel deep learning-based method to predict an optimized structure for a given boundary condition and optimization setting without using any iterative scheme. For this purpose, first, using open-source topology optimization code, datasets of the optimized structures paired with the corresponding information on boundary conditions and optimization settings are generated at low (32 × 32) and high (128 × 128) resolutions. To construct the artificial neural network for the proposed method, a convolutional neural network (CNN)-based encoder and decoder network is trained using the training dataset generated at low resolution. Then, as a two-stage refinement, the conditional generative adversarial network (cGAN) is trained with the optimized structures paired at both low and high resolutions, and is connected to the trained CNN-based encoder and decoder network. The performance evaluation results of the integrated network demonstrate that the proposed method can determine a near-optimal structure in terms of pixel values and compliance with negligible computational time.

## 1. Introduction

Recent rapid advances in deep learning algorithms and computer hardware technology have led to successful applications of machine learning in various fields, including medical diagnosis, e-commerce, speech recognition, and health monitoring. Deep neural networks are flexible in that they allow "deep" learning of a high-order function for a target model based on the training dataset. In particular, several studies have explored the potential of applying deep learning to computational physics and engineering fields, such as quantum and particle physics as well as fluid dynamics simulation (Carleo and Troyer 2017; Tompson et al. 2017; Singh et al. 2017; Mills et al. 2017) In regard to structural analysis, Lee *et al*. (2017) recently used deep learning to predict the structural behavior of simple truss structures without conducting finite element analysis.

Among various methods used in deep learning, generative models, which aim at learning the true data distribution of input data, are promising for applications in the fields of computational physics and engineering. In general, generative models have been used to classify target objects (e.g., classification of handwritten digits (Lecun et al. 1998)) as well as generate realistic fake images with limited information (e.g., drawing artificial faces of women with eyeglasses (Radford et al. 2016)). Farimani *et al*. (2017) recently developed a data-driven paradigm to model and simulate the physics of transport phenomena using a generative adversarial network (GAN) (Goodfellow et al. 2014). In addition, Pagnini *et al*. (2018) also used the GAN to simulate three-dimensional (3D) high-energy particle showers in electromagnetic calorimeters. Wetzel (2017) applied a variational autoencoder (Kingma and Welling 2014) to describe the states of the two-dimensional Ising model.

As a branch of design optimization, topology optimization (O. Sigmund 2003) is a systematic method of determining the optimal distribution of materials in the design space under a given boundary condition in order to extremize the system performance while satisfying the design requirements (or constraints). Topology optimization offers a distinctive benefit of determining a feasible conceptual design; however, it has an inherent drawback in terms of computational cost because of a large number of design variables and/or iterations during optimization. To date, there have been many studies to reduce computational cost incurred in topology optimization. Kim and Yoon (2000) proposed a multi-level strategy in which topology optimization is progressively performed from low to high resolution. Jang and Kwak (2006, 2007) proposed a design space optimization method that involves conducting a series of design space adjustments and refinements during optimization. Kim *et al.* (2012) proposed a new convergence criterion to reduce computational expense by adaptively reducing the number of design variables on the basis of the optimization history of each design variable. Wu *et al.* (2016) proposed a new method of performing topology optimization with a high-performance GPU solver. As alternative approaches, several studies have recently attempted to enhance the computational efficiency of topology optimization by adopting machine learning. In particular, Liu *et al.* (2015) reduced the dimensionality of design variables using K-means clustering, and performed metamodeling-based topology optimization. Aulig and Olhofer (2014) proposed a method of substituting analytical sensitivities with those approximated by a trained artificial neural network. Sosnovik and Oseledets (2017) adopted a deep learning-based image segmentation task to hasten topology optimization; their model learned about the mapping from an intermediate structure at each iteration to obtain the final optimized structure. However, their neural network did not consider information on boundary conditions and optimization settings that are essential to perform topology optimization.

To the best of our knowledge, this is the first time a deep learning-based method is proposed to determine a near-optimal topological design without any iteration. First, supervised learning is performed to learn a target function for optimal material distribution from a training dataset, which includes information on boundary conditions and optimization settings paired with the corresponding optimized structures. Then, a two-stage refinement is performed to efficiently predict a more refined structure by implementing GANs. Through verification with a test dataset, the proposed method was shown to be feasible for determining a near-optimal topological design without any time-consuming iterative process.

## 2. Overview of machine learning and deep learning
### 2.1 Artificial neural network

In 1959, the term "machine learning" was first introduced by Arthur Samuel who is one of the pioneers in artificial intelligence; this terminology literally means the ability to learn without being explicitly programmed. Ideally, a machine that has been successfully trained using a dataset can predict, judge, and/or reason the results of new input data. Deep learning is a part of machine learning (Fig. 1), and stems from artificial neural networks (ANNs), which are inspired by the biological neural network of the brain.

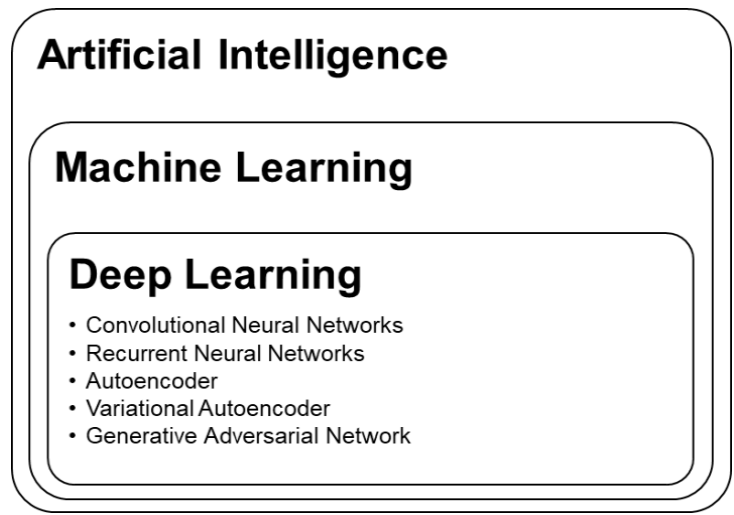

**Fig. 1.** Machine learning and deep learning

In principle, an ANN consists of a collection of nodes called artificial neurons. These neurons can receive multiple inputs and then calculate an output value though an activation function, as shown in Fig. 2.

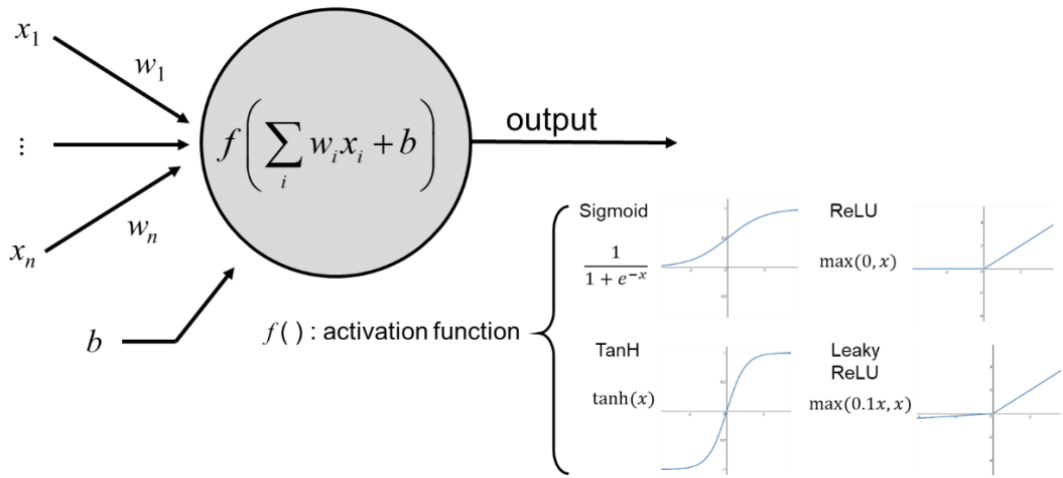

(a) Artificial neurons

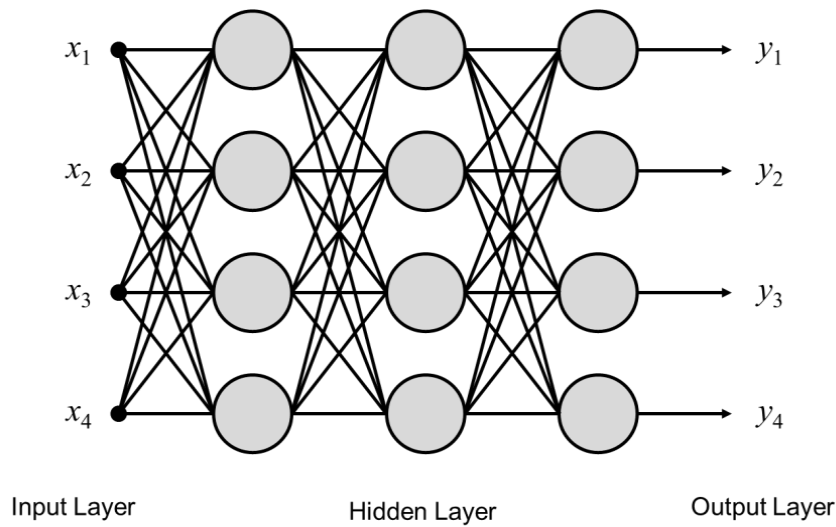

(b) Artificial neural networks

**Fig. 2.** Schematic of an artificial neural network

Through the training process with the prepared dataset, optimization algorithm determines the optimal weights of an ANN (i.e., $w_i$ and $b$ in Fig. 2(a)) that can minimize a loss function (e.g., prediction error). Although the concept of the ANN was originally introduced in the 1960s, it had been impossible to train a large-scale ANN until the 2000s. The advent of a new activation function called the rectified linear unit (ReLU) (Hahnioser et al. 2000; Nair and Hinton 2010) and effective weight initialization (Hinton et al. 2006) allowed for rapidly processing a considerable amount of data using enhanced hardware such as general-purpose computing on graphics processing units (GPGPU). The introduction of open-source deep learning frameworks such as Theano (Bergstra et al. 2010), Tensorflow (Abadi et al. 2015), and

Keras (Chollet 2015) also promoted the application of deep learning to other fields. In this study, the Keras framework with a TensorFlow backend was used as the application programming interface (API) to model and train the ANN.

Depending on the different tasks that an ANN is used for, the most common modifications to the ANN include the convolutional neural network (CNN), which is suitable for image recognition; recurrent neural network (RNN), which is suitable for time series data; and autoencoder for dimension reduction.

## 2.2 Convolutional neural network

Convolution operation, which has been widely used in signal processing and image recognition, can be used to extract features from given data. In typical image recognition, features are extracted using a pre-defined filter and are then classified into the given labels. Therefore, the performance of convolution operation significantly depends on the filter selected. Conversely, in the CNN, the parameters of a filter are trained along with the training of the classifier. LeCun *et al.* (1989; 1998) proposed this method to classify the MNIST database, which includes handwritten digits. DeepMind's AlphaGo (Silver et al. 2016) also implemented CNN to learn how to play Go.

Convolution operation in ANNs is a cross-correlation between input data and a convolutional filter, which can be represented in terms of a matrix of weights. As shown in Fig. 3, a convolution filter strides on the input image to produce a feature map as output. A stride denotes the amount by which the filter moves at each step.

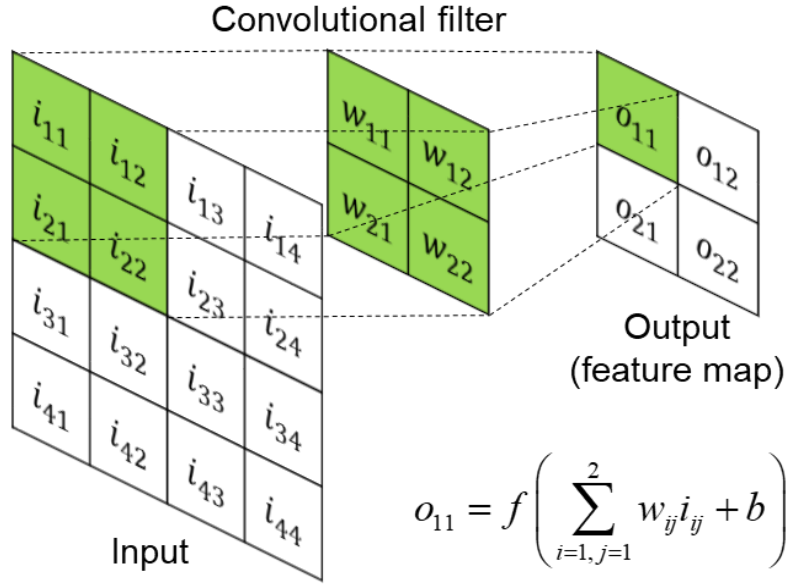

**Fig. 3.** Example of a convolutional neural network with a stride of 2

**2.3 Generative model and generative adversarial network**

In machine learning, a generative model is used to approximate a probability density function of given data. For example, in order to create a new face image, it is essential to first estimate the probability density distribution of the face images of humans. The objective of generative models is to find a probability model *P*(*x*) that can represent the distribution in the input data space *x* and generate new data based on the trained model. The outstanding characteristic of generative models is the utilization of the latent variable space. Here, latent variables denote attribute variables that can be used to reduce dimensionality of input data space.

Among various generative models, the GAN was first proposed by Goodfellow *et al.* (2014), and in recent times, has become one of the most vibrant research topics in machine learning. Taking advantage of concepts in game theory, the GAN architecture consists of a generator and discriminator, both of which compete with each other in the training phase. As depicted in Fig. 4, the generator *G*(**z**) produces fake data from the latent random variable **z**. The discriminator *D*(**y**) takes either real or fake data as input and returns the probability that the given input **y** is real (Fig. 4). The generator *G* is trained to generate realistic fake outputs that the discriminator *D* cannot distinguish from real data, whereas the discriminator $D$ is trained to precisely distinguish between fake and real data. Thus, the generator $G$ can be formulated to minimize

$$L^{GAN} = E_{y \sim p_{data}(\mathbf{y})}[\log(D(\mathbf{y}))] + E_{\mathbf{z} \sim p(\mathbf{z})}[\log(1-D(G(\mathbf{z})))] \tag{1}$$

where $p_{data}(\mathbf{y})$ and $p(\mathbf{z})$ are the probability density functions of real data $\mathbf{y}$ and latent variable $\mathbf{z}$, respectively. Conversely, the discriminator $D$ needs to be formulated to maximize the same loss function. GAN has been widely used to generate realistic images (Radford et al. 2016; Karras et al. 2018), audio (Wang et al. 2017), and text (Fedus et al. 2018).

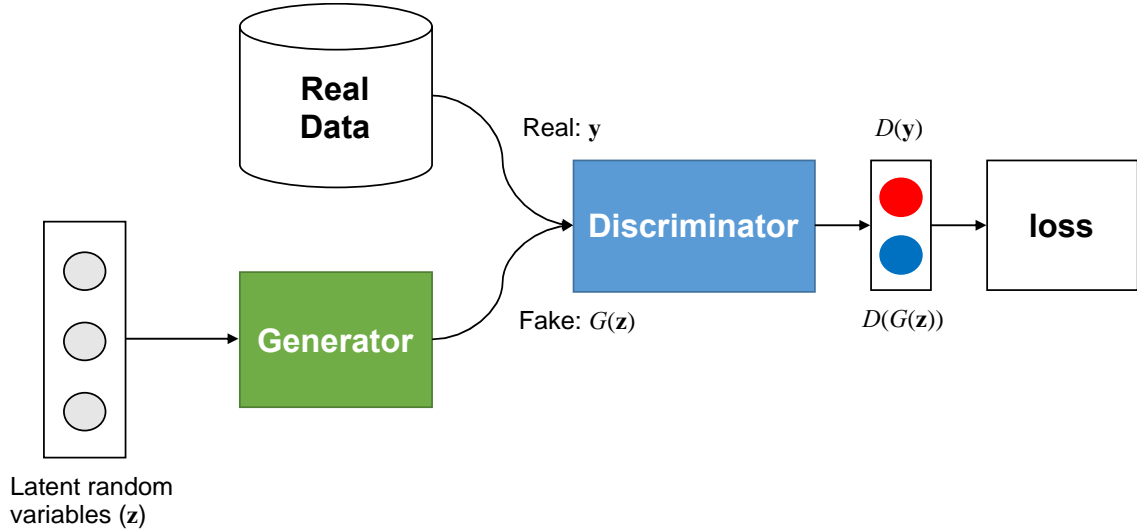

**Fig. 4.** Conceptual flow of a generative adversarial network

## 3. Proposed Method

The conceptual flow of the proposed method is shown in Fig. 5. First, datasets of the optimized structures paired with the information on optimization settings and boundary conditions were generated using the open-source topology optimization code (Andreassen et al. 2011). Each dataset was created at both low (32 × 32) and high (128 × 128) resolutions under the same boundary conditions and optimization settings. These datasets were divided into training dataset for the training (or learning) process and test dataset for performance evaluation. The CNN-based encoder and decoder network was trained with the training dataset generated at low resolution. Then, for progressive refinement, the conditional GAN was trained with the paired optimized structures at both low and high resolutions. The performance of the integrated network that connects the trained network of Step 2 and that of Step 3 was evaluated using the test dataset.

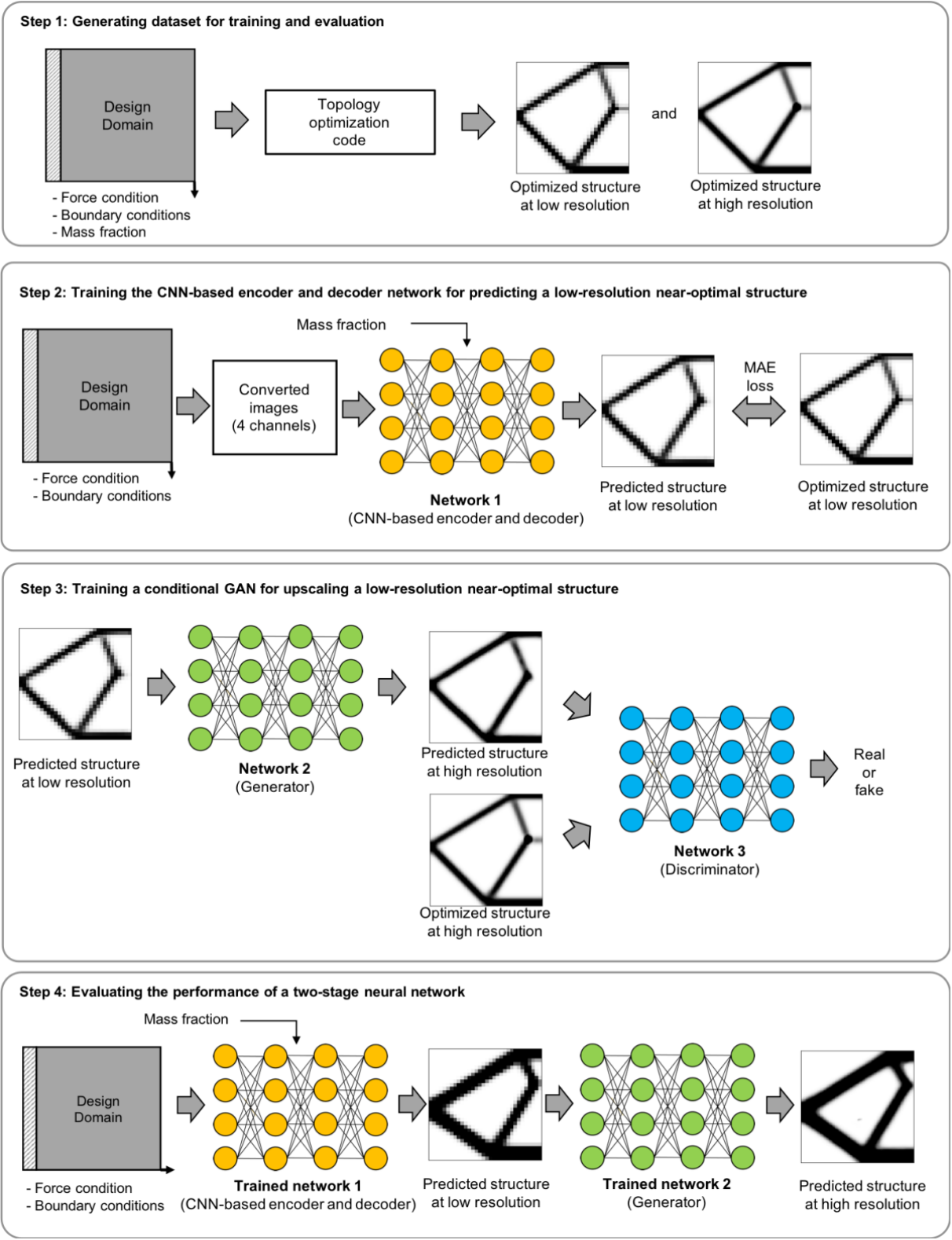

**Fig. 5.** Conceptual flow of the proposed method

### 3.1 Step 1: Generating the dataset for training and evaluation

For an appropriate training process, major information used in topology optimization (e.g., force and geometric boundary conditions, passive zone, and mass fraction) needs to be encoded into latent variables. This information can be generally categorized as follows:

- Nodal Condition: Force and geometric boundary condition, etc.
- Elementwise Condition: Passive zone, which is predetermined as void or solid, etc.
- Scalar Condition: Mass fraction, filtering radius, etc.

As the first step, in this study, we considered a single concentrated force, geometric boundary condition on the side, and mass fraction. The design domain was fixed to be 32 × 32 grids at low resolution and 128 × 128 grids at high resolution. A total of 100,000 optimized structures were generated as the dataset by repeatedly executing the open-source topology optimization code (Andreassen et al. 2011) with the randomized conditions as follows:

- Mass fraction: 0.2 to 0.8
- Application point of single concentrated force: Node 1 to 1089
- Angle of force: 0° to 360°
- Fixed boundary condition

The output of the ANN (i.e., a predicted near-optimal design at low resolution) can be directly represented as a matrix of relative densities ranging from 0 to 1. However, the input data for boundary conditions and optimization settings need to be transformed into a proper form for training. In this study, force and geometric boundary conditions were transformed into images with four input channels (Fig. 6). At low resolution, 32 × 32 matrices and 33 × 33 matrices were used to represent the optimized structures and boundary conditions, respectively. All input channels were initialized to zero. The first and second input channels represent the force boundary conditions on the x- and y-axes, respectively. The magnitude of the corresponding force was inputted to the grid value of each channel. Further, the third and fourth input channels represent the geometric boundary conditions. A value of 1 in the grid indicates that a fixed boundary condition is imposed on the corresponding node.

The average time to generate a single dataset using the topology optimization code was 0.379 s and 23.4 s at low and high resolutions, respectively. The entire dataset was divided into training, validation and test datasets with a 64:16:20 ratio. All the calculations for training and evaluation were performed on the Google Cloud Platform with the following specification:

- CPU : Intel Xeon (8 cores and 2.5 GHz CPU clock)
- GPU : NVIDIA Tesla P100 (3584 CUDA cores of 1.19 GHz GPU clock and 16 GB memory)

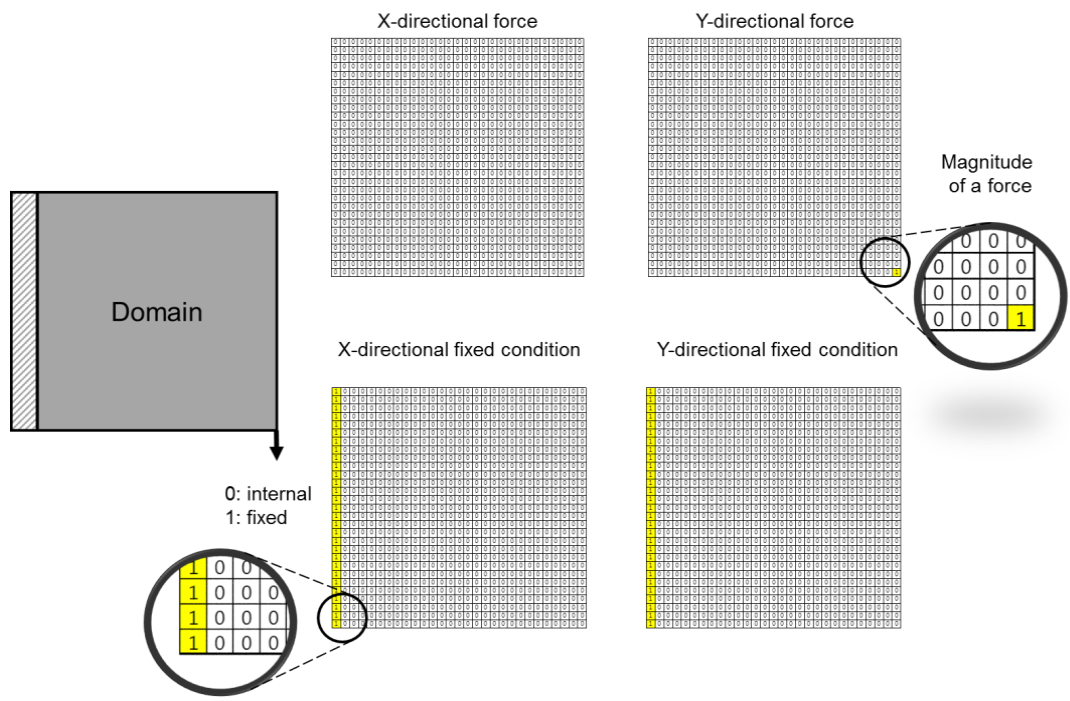

**Fig. 6.** Discretization of force and geometric boundary conditions

### 3.2 Step 2: Training the CNN-based encoder and decoder network

In this study, instead of using a fully connected network, a two-dimensional CNN was used to implement the encoder and decoder network. Figure 7 presents the detailed encoder and decoder network for supervised learning. In the encoder network, boundary conditions are encoded into latent variables through the seven CNN (red arrows in Fig. 7) and three max-pooling layers (blue arrows in Fig. 7). The decoder network, which is a mirror function of the encoder, consists of the nine CNN (red and yellow arrows in Fig. 7) and three upscaling layers (purple arrows in Fig. 7) in order to generate a structure using the latent variables. Mass fraction, which is a scalar value and does not represent spatial information, was directly inputted to the layer of latent variables. The encoder and decoder network except the output layer used a ReLU activation (Nair and Hinton 2010), while the output layer used a sigmoid function (Mitchell 1997), which can directly handle outputs ranging from 0 to 1. The reconstructed loss was calculated in terms of the mean absolute error (MAE) between the optimized (i.e., obtained from topology optimization) and predicted (i.e., obtained using the proposed method) structures. Considering computational costs and accuracy of the test dataset, the detailed parameters (number of layers, features maps, filter size, etc.) were determined with trial and error.

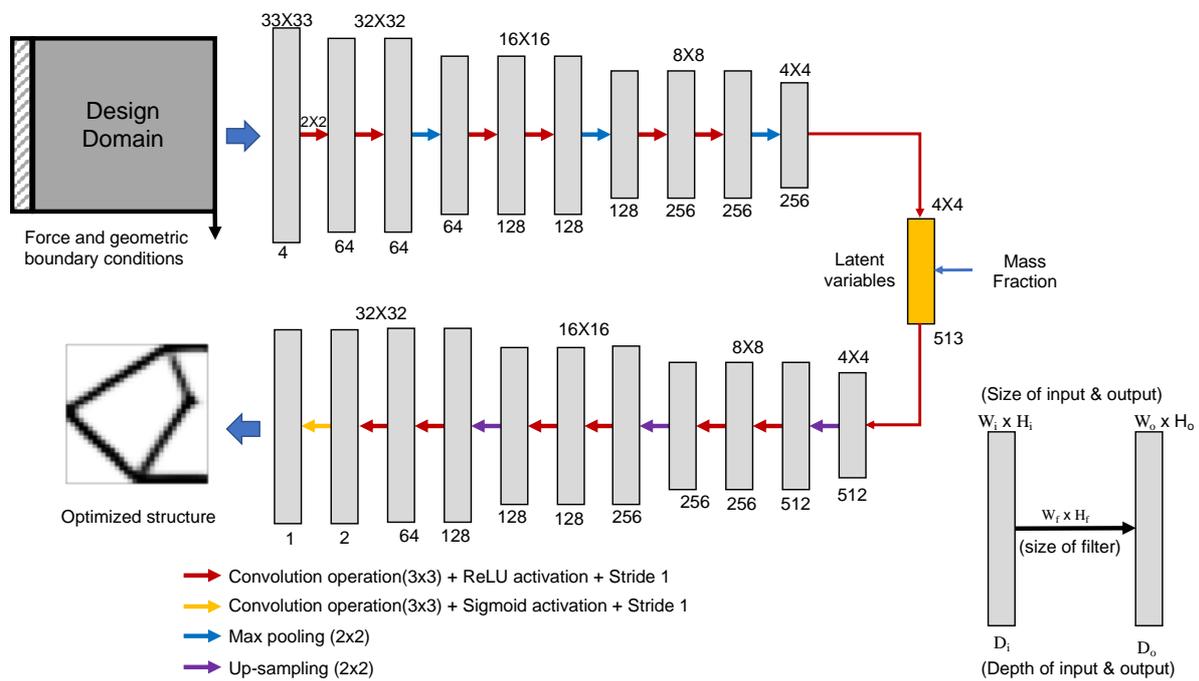

**Fig. 7.** Architecture of the encoder and decoder networks for learning the topology optimization dataset

For the training process, the ADAM optimizer (Kingma and Ba 2014), which is a gradient-based optimization algorithm for stochastic objective functions, was used to determine the optimal weights of the encoder and decoder network, which minimize the loss between input and output. Because the calculation of a loss function for the entire dataset incurs high computational cost, the dataset is generally divided into a set of mini-batches for one epoch. In machine learning, one epoch represents an entire dataset passing through the ANN once. In this study, the batch size, which is the number of datasets in a single mini-batch, was set to be 256, considering the size of GPU memory and required training accuracy. Figure 8 shows the history of training and validation loss. The training process was terminated when the validation loss reached the minimum value; after 122 epochs, the MAEs of training and validation datasets were 0.99% and 2.20%, respectively.

In the next section, we discuss the two-stage refinement process, which is a recently introduced concept in deep learning (Zhang et al. 2017; Karras et al. 2018); it will be used to more efficiently determine a near-optimal structure of higher-resolution image.

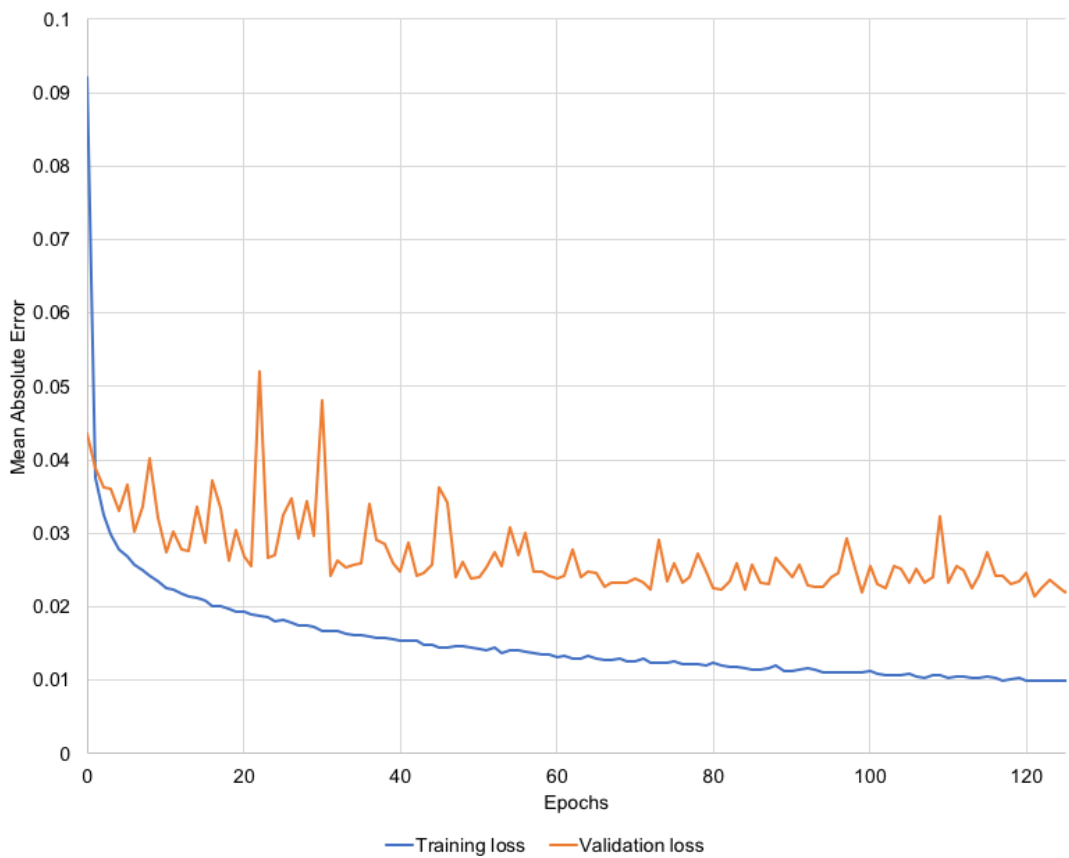

**Fig. 8.** Loss curves for the training and validation datasets in Step 2

**3.3 Step 3: Training a conditional GAN to upscale a low-resolution near-optimal structure**

Conditional GAN (cGAN), which is a variation of GAN, can handle additional conditional variables in addition to or in place of the samples for the generator and discriminator (Mirza and Osindero 2014), as follows:

$$G:(\mathbf{c},\mathbf{z}) \rightarrow y_g \quad and \quad D:(\mathbf{c},\mathbf{y}) \rightarrow p_g, \qquad (2)$$

where $\mathbf{c}$ is a conditional variable vector, $\mathbf{z}$ is a latent random variable vector, $\mathbf{y}_g$ is the generated (fake) output, and $p_g$ is the probability that $\mathbf{y}$ is real. In this study, the conditional variables ($\mathbf{c}$ in Eq. (2)) were the optimized structures of low resolution, and the generated output ($\mathbf{y}_g$ in Eq. (2)) was the predicted structures of high resolution. Latent random variables were not considered because the conditional variables can be used to generate the output themselves. Therefore, the loss function, $L^{cGAN}$, can be expressed as

$$L^{cGAN} = E_{\mathbf{c},\mathbf{y} \sim p_{data}(\mathbf{c},\mathbf{y})}[\log(D(\mathbf{c},\mathbf{y}))] + E_{\mathbf{c},\mathbf{y} \sim p_{data}(\mathbf{c},\mathbf{y})}[\log(1-D(\mathbf{c},G(\mathbf{c})))] \qquad (3)$$

where $\mathbf{c}$ is the optimized structure of low resolution. After training the two adversarial networks ($G$ and $D$) with $L^{cGAN}$, $G$ can generate a near-optimal structure of high resolution, which is predicted from the optimized structure of low resolution, and $D$ discriminates the optimized (real) and predicted (fake) structures.

Figure 9 shows the architecture of the cGAN used for upscaling in our study. In Fig. 9, for clarity, a residual CNN block (He et al. 2016) is represented using yellow arrows. Further, gray boxes indicate two-dimensional layers for images or feature maps, while white boxes indicate one-dimensional layers. Taking the optimized structures of low resolution as input, the generator was trained to predict the optimal structure of high resolution as realistically as possible. At the same time, using the real dataset (predicted structures of low resolution and optimized structure of high resolution) and fake dataset (predicted structures of both low and high resolution), the discriminator was trained to distinguish between the real (optimized) and fake (predicted) dataset. To elaborate, the deep residual network (He et al. 2016), in which the identity skip-connection alleviates the vanishing gradient problem, was implemented for the generator network. The generator takes the input images of $32 \times 32$ resolution and performs convolution operations with a single convolutional layer and eight residual CNN blocks. Finally, two consecutive 2x upscaling layers and convolution with a filter size of $5 \times 5$ generate the output images of $128 \times 128$ resolution.

The discriminator of the proposed cGAN model takes two different types of inputs: the predicted structures of low resolution and optimized structures of high resolution for real data; and the predicted structures of low resolution and predicted structures of high resolution for fake data. In particular, the $32 \times 32$ resolution input is upscaled to $128 \times 128$ and then is paired with the corresponding optimized (real) or predicted (fake) structure of $128 \times 128$ resolution as two channel images. Then, the combined input data passes through a series of convolutional layers as well as a fully connected layer, as shown in Fig. 9. The source code of the proposed method is based on the SRGAN code (Github; Ledig et al. 2016), which demonstrates the highest level of performance for single image super-resolution.

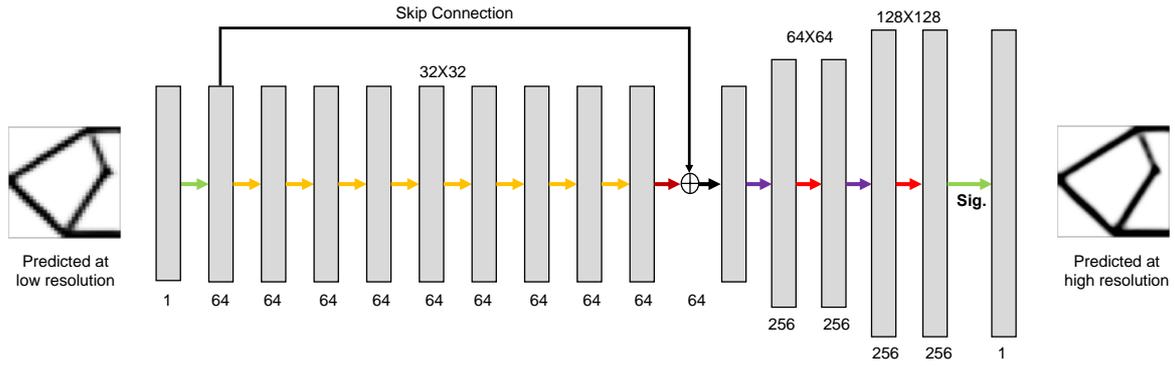

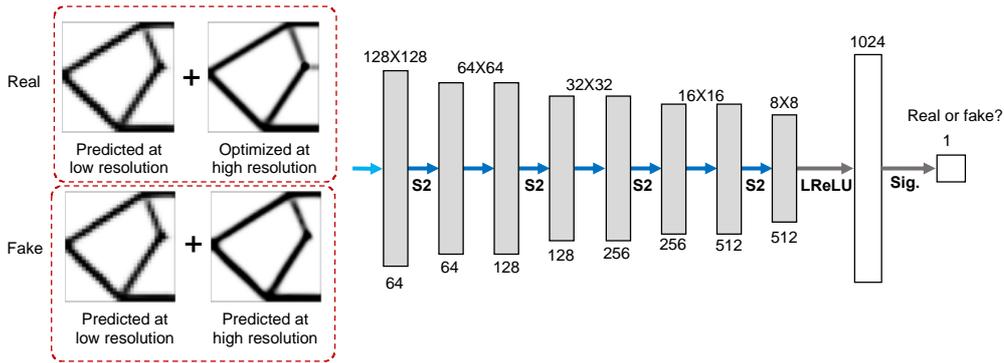

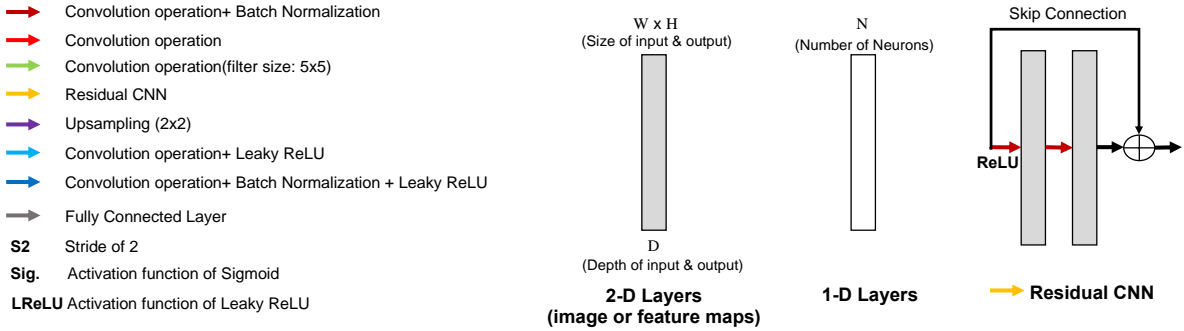

**Fig. 9** Architecture of the conditional GAN for upscaling topology optimization results (Here, a stride of 1, a filter size of 3 × 3, and activation function of ReLU were used unless indicated otherwise.)

To properly train the generator, the MAE was added to the loss function of cGAN. Thus, the final objective function of the proposed model can be expressed as

$$L^{final} = L^{cGAN} + \lambda E_{\mathbf{c},\mathbf{y} \sim p_{data}(\mathbf{c},\mathbf{y})}[\|\mathbf{y} - G(\mathbf{c})\|_1] \tag{4}$$

where $\lambda$ is a weight of MAE loss. The generator and discriminator were alternatively trained with 64,000

paired data using a batch size of 16. The value of $\lambda$ was selected to be 0.001, and the ADAM optimizer (Kingma and Ba 2014) was used. After training process of 20 epochs, the MAEs of training and validation datasets were 1.98% and 2.74%, respectively.

Through the training process of two adversarial networks, the networks can "understand" the optimal material distribution in the training data. The discriminator is trained to discriminate a difference between the optimized and predicted structures and, simultaneously, the generator is trained to predict a realistic optimal structure so that the discriminator cannot discriminate a difference. Therefore, more realistic structures can be generated as the training process progresses further.

**3.4 Step 4: Evaluating the performance of the integrated network**

As shown in Fig. 5, the output layer of the CNN-based encoder and decoder network used in Step 2 was connected to the input layer of the generator network used in Step 3 in order to efficiently determine a near-optimal structure of high resolution. The performance of the integrated network was evaluated using the test dataset. Figure 10 shows the predicted structures of low resolution (first column) and high resolution (second column) for a given boundary conditions and mass fraction; in addition, it shows the corresponding optimized structures of low resolution (third column) and high resolution (fourth column). It is clearly demonstrated that the predicted structures of high resolution are visually similar to the optimized structures of high resolution. In addition, Tables 1 and 2 show the performance of the proposed method in terms of computational time and prediction errors. Because the proposed method can determine a near-optimal structure without any iteration, the computational time of the proposed method is only 0.06% of that of topology optimization used at the same $128 \times 128$ resolution. In terms of accuracy, the pixel-wise MAE and mass fraction of the predicted structures of high resolution are less than 2.72% and 0.51%, respectively; these values indicate that the proposed method could successfully determine a "near-optimal" structure. However, as shown in Fig. 11, the errors of compliance in some cases (specifically, 4.19% of the test data in this study) became significantly large when structural discontinuity occurred; the reason might be that no measure of structural strength (e.g., compliance) was directly evaluated and used in the proposed neural network. It should be noted that two structures that are visually similar to each other can have different compliances. With further work, this error could be reduced by implementing additional networks that can directly consider structural connectivity and/or structural strength.

Table 1. Comparison of computational time between topology optimization and the proposed method

|  | Topology optimization | Proposed method |
|---|---|---|
| Average computational time | 22.7 s | 0.014 s (0.06% of that of topology optimization) |

Table 2. Comparison of accuracy between the optimized and predicted structures

|  | Pixel values | Compliance | Mass fraction |
|---|---|---|---|
| Average error | 2.72% | 4.19% for normal cases (96.0% of test data) / 21204300% for disconnected cases (4.0% of test data) | 0.51% |

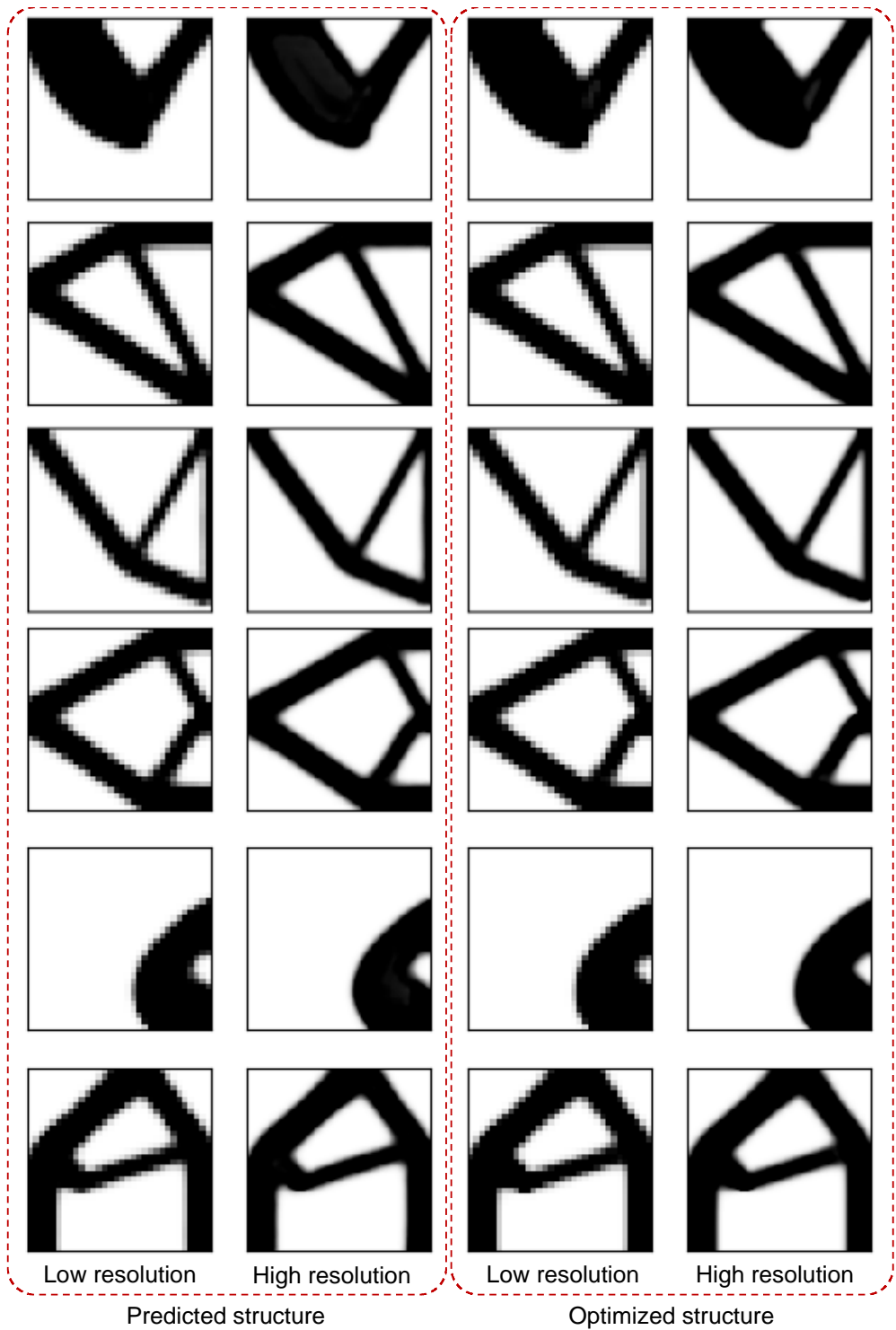

**Fig. 10** Comparison of the predicted structures of low resolution (first column) and high resolution (second column) for given boundary conditions and mass fraction, and the corresponding optimized structures of low resolution (third column) and high resolution (fourth column)

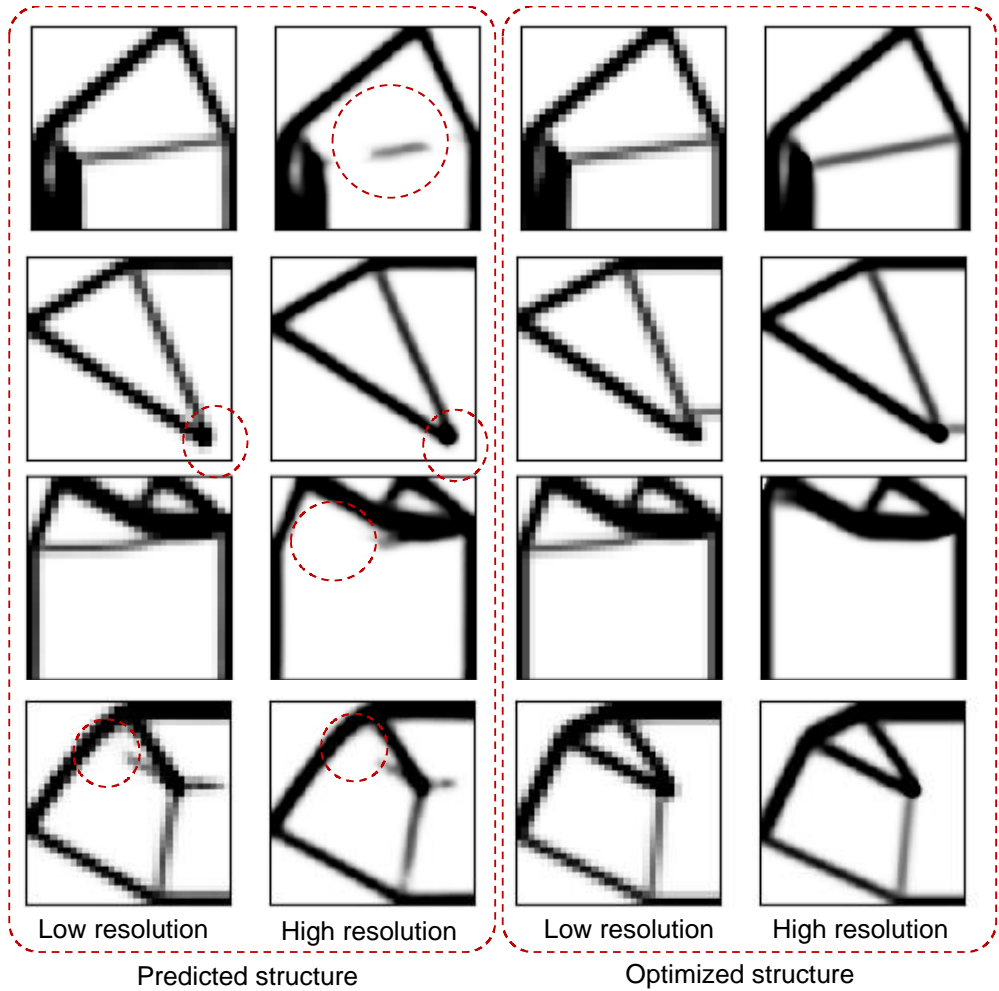

**Fig. 11** Examples of the predicted structures with structural disconnection

## 4. Conclusions

In this paper, a deep learning method was implemented to determine a near-optimal topological design for the given boundary conditions and optimization settings without any iteration. For dataset preparation, the boundary conditions were transformed into multi-channel images paired with their optimized structures. These images were encoded into latent variables through the CNN-based encoder network, and latent variables in the decoder network were then used to reconstruct a near-optimal structure. For progressive refinement, cGAN was connected to the trained CNN-based encoder network to efficiently determine a near-optimal structure of higher resolution. Numerical examples demonstrate that the proposed method can determine a near-optimal structure in terms of pixel values and compliance with negligible computational cost.

However, several limitations should be addressed for further real-world applications. First, as the size of the finite element model increases, it becomes more difficult to prepare a sufficiently large training dataset to train a larger ANN. It is worthy to note that the state-of-the-art GAN can handle the maximum output size of $1024 \times 1024$ and $64 \times 64 \times 64$ for two- and three-dimensional cases, respectively (Liu et al. 2017; Karras et al. 2018). Second, as the first step toward a deep learning-based design method, only simplified, limited boundary conditions and optimization setting were considered in this study. To consider more general conditions, more datasets and larger ANNs would be required. This will be undertaken as part of our future work. Finally, this study was limited in its handling of a fixed and regular mesh. However, this limitation could be overcome because recent studies have considered non-Euclidean meshes in deep learning (Defferrard et al. 2016; Kipf and Welling 2016; Qi et al. 2017).

Despite the aforementioned limitations of the proposed method in its current form, it is pioneering and intriguing to incorporate deep learning with topology optimization. If the proposed method is combined with conventional topology optimization, the convergence of the optimization process can be significantly improved. For example, a predicted structure can be set as the initial design of topology optimization. With further work, topology optimization with reduced latent variables could also drastically boost the convergence and computational burden of topology optimization.

**Acknowledgments**

This research was supported by the National Research Foundation of Korea (NRF) grant funded by the Korea government (MSIT) (NRF-2018R1C1B6005157), and National Institute of Supercomputing and Network(NISN)/Korea Institute of Science and Technology Information(KISTI) with supercomputing resources including technical support (KSC-2017-S1-0029).

**References**

Abadi M, Agarwal A, Paul Barham EB, et al (2015) TensorFlow: Large-scale machine learning on heterogeneous systems. https://www.tensorflow.org/.

Andreassen E, Clausen A, Schevenels M, et al (2011) Efficient topology optimization in MATLAB

using 88 lines of code. Struct Multidiscip Optim 43:1–16. doi: 10.1007/s00158-010-0594-7

Aulig N, Olhofer M (2014) Topology optimization by predicting sensitivities based on local state features. 11th World Congr Comput Mech (WCCM XI) 437–448. doi: 10.1145/2576768.2598314

Bergstra J, Breuleux O, Bastien FF, et al (2010) Theano: a CPU and GPU math compiler in Python. Proc Python Sci Comput Conf.

Carleo G, Troyer M (2017) Solving the quantum many-body problem with artificial neural networks. Science (80- ) 355:602–606. doi: 10.1126/science.aag2302

Chollet F and others (2015) Keras. In: GitHub Repos. https://github.com/fchollet/keras. Accessed 17 Nov 2017

Defferrard M, Bresson X, Vandergheynst P (2016) Convolutional neural networks on graphs with fast localized spectral filtering. Conf Neural Inf Process Syst.

Farimani AB, Gomes J, Pande VS (2017) Deep learning the physics of transport phenomena. ArXiv: 1709.02432.

Fedus W, Goodfellow I, Dai AM (2018) MaskGAN: Better text generation via filling in the______. ArXiv: 1801.07736.

Github: Keras GAN/srgan. GitHub Repos https://github.com/eriklindernoren/Keras-GAN/blob/.

Goodfellow IJ, Pouget-Abadie J, Mirza M, et al (2014) Generative adversarial networks. Conf Neural Inf Process Syst. doi: 10.1001/jamainternmed.2016.8245

Hahnioser RHR, Sarpeshkar R, Mahowald MA, et al (2000) Digital selection and analogue amplification coexist in a cortex- inspired silicon circuit. Nature 405:947–951. doi: 10.1038/35016072

He K, Zhang X, Ren S, Sun J (2016) Deep residual learning for image recognition. IEEE Conf Comput Vis Pattern Recognit 770–778. doi: 10.1109/CVPR.2016.90

Hinton GE, Osindero S, Teh Y-W (2006) A fast learning algorithm for deep belief bets. Neural Comput 18:1527–1554. doi: 10.1162/neco.2006.18.7.1527

Jang IG, Kwak BM (2006) Evolutionary topology optimization using design space adjustment based on fixed grid. Int J Numer Methods Eng 66:1817–1840. doi: 10.1002/nme.1607

Jang IG, Kwak BM (2007) Design space optimization using design space adjustment and refinement. Struct Multidiscip Optim 35:41–54. doi: 10.1007/s00158-007-0112-8

Karras T, Aila T, Laine S, Lehtinen J (2018) Progressive growing of GANs for improved quality, stability, and variation. Int Conf Learn Represent. doi: 10.1002/joe.20070

Kim SY, Kim IY, Mechefske CK (2012) A new efficient convergence criterion for reducing computational expense in topology optimization: reducible design variable method. Int J Numer Methods Eng 90:752–783. doi: 10.1002/nme.3343

Kim YY, Yoon GH (2000) Multi-resolution multi-scale topology optimization — a new paradigm. Int J Solids Struct 37:5529–5559. doi: 10.1016/S0020-7683(99)00251-6

Kingma DP, Ba J (2014) Adam: A method for stochastic optimization. Int Conf Learn Represent. doi: 10.1145/1830483.1830503

Kingma DP, Welling M (2014) Auto-encoding variational bayes. Int Conf Learn Represent. doi: 10.1051/0004-6361/201527329

Kipf TN, Welling M (2016) Semi-supervised classification with graph convolutional networks. Int Conf Learn Represent. doi: 10.1051/0004-6361/201527329

LeCun Y, Boser B, Denker JS, et al (1989) Backpropagation applied to handwritten zip code recognition. Neural Comput 1:541–551. doi: 10.1162/neco.1989.1.4.541

Lecun Y, Bottou L, Bengio Y, Haffner P (1998) Gradient-based learning applied to document recognition. Proc IEEE 86:2278–2324. doi: 10.1109/5.726791

Ledig C, Theis L, Huszar F, et al (2016) Photo-realistic single image super-resolution using a generative adversarial network. doi: 10.1109/CVPR.2017.19

Lee S, Ha J, Zokhirova M, et al (2017) Background information of deep learning for structural engineering. Arch. Comput. Methods Eng. 0:1–9.

Liu J, Yu F, Funkhouser T (2017) Interactive 3D modeling with a generative adversarial network. doi: 10.1109/3dv.2017.00024

Liu K, Tovar A, Nutwell E, Detwiler D (2015) Towards nonlinear multimaterial topology optimization using unsupervised machine learning and metamodel-based optimization. Des Autom Conf. doi: 10.1115/DETC2015-46534

Mills K, Spanner M, Tamblyn I (2017) Deep learning and the Schrödinger equation. Phys Rev A 96:042113. doi: 10.1103/PhysRevA.96.042113

Mirza M, Osindero S (2014) Conditional generative adversarial Nets. ArXiv: 1411.1784.

Mitchell TM (1997) Machine Learning. McGraw-Hill, pp 96–97

Nair V, Hinton GE (2010) Rectified linear units improve restricted boltzmann machines. Proc 27th Int Conf Mach Learn 807–814. doi: 10.1.1.165.6419

O. Sigmund MPB (2003) Topology optimization theory, methods, and Applications. Springer

Paganini M, de Oliveira L, Nachman B (2018) CaloGAN: Simulating 3D high energy particle showers in multilayer electromagnetic calorimeters with generative adversarial networks. Phys Rev D 97:014021. doi: 10.1103/PhysRevD.97.014021

Qi CR, Su H, Mo K, Guibas LJ (2017) PointNet: Deep learning on point sets for 3D classification and segmentation.

Radford A, Metz L, Chintala S (2016) Unsupervised representation learning with deep covolutional generative adversarial networks. Int Conf Learn Represent. doi: 10.1051/0004-6361/201527329

Silver D, Huang A, Maddison CJ, et al (2016) Mastering the game of Go with deep neural networks and tree search. Nature 529:484–489. doi: 10.1038/nature16961

Singh AP, Medida S, Duraisamy K (2017) Machine-learning-augmented predictive modeling of turbulent separated flows over airfoils. AIAA J 55:2215–2227. doi: 10.2514/1.J055595

Sosnovik I, Oseledets I (2017) Neural networks for topology optimization. ArXiv: 1709.09578

Tompson J, Schlachter K, Sprechmann P, Perlin K (2017) Accelerating eulerian fluid simulation with convolutional networks. Proc Mach Learn Res 70:3424–3433.

Wang Y, Skerry-Ryan RJ, Stanton D, et al (2017) Tacotron: Towards end-To-end speech synthesis. Proc Annu Conf Int Speech Commun Assoc INTERSPEECH 2017–Augus:4006–4010. doi: 10.21437/Interspeech.2017-1452

Wetzel SJ (2017) Unsupervised learning of phase transitions: From principal component analysis to variational autoencoders. Phys Rev E 96:1–8. doi: 10.1103/PhysRevE.96.022140

Wu J, Dick C, Westermann R (2016) A System for high-resolution topology optimization. IEEE Trans Vis Comput Graph 22:1195–1208. doi: 10.1109/TVCG.2015.2502588

Zhang H, Xu T, Li H, et al (2017) StackGAN++: Realistic image synthesis with stacked generative adversarial networks. doi: 10.1109/ICCV.2017.629